\documentclass[letterpaper, 10 pt, conference]{ieeeconf}  % Comment this line out if you need a4paper

\IEEEoverridecommandlockouts                              % This command is only needed if 
\usepackage{booktabs}
\usepackage{graphicx}
\usepackage{cuted}
\usepackage{capt-of}
\usepackage{amsmath}
\usepackage{amssymb}

\title{\LARGE \bf
QTOP: Scaling Humanoid Motion Tracking with Quantile Critics}

\author{
\authorblockN{Markus Peschl, Dana Kianfar, Daniel Dijkman}
\authorblockA{Qualcomm AI Research\authorrefmark{1}\\
}
\thanks{\authorrefmark{1}Qualcomm AI Research is an initiative of Qualcomm Technologies, Inc.}
}

\begin{document}

\maketitle
\thispagestyle{empty}
\pagestyle{empty}

%%%%%%%%%%%%%%%%%%%%%%%%%%%%%%%%%%%%%%%%%%%%%%%%%%%%%%%%%%%%%%%%%%%%%%%%%%%%%%%%

\begin{strip}
\vspace{-50pt}
\centering
\includegraphics[width=\textwidth]{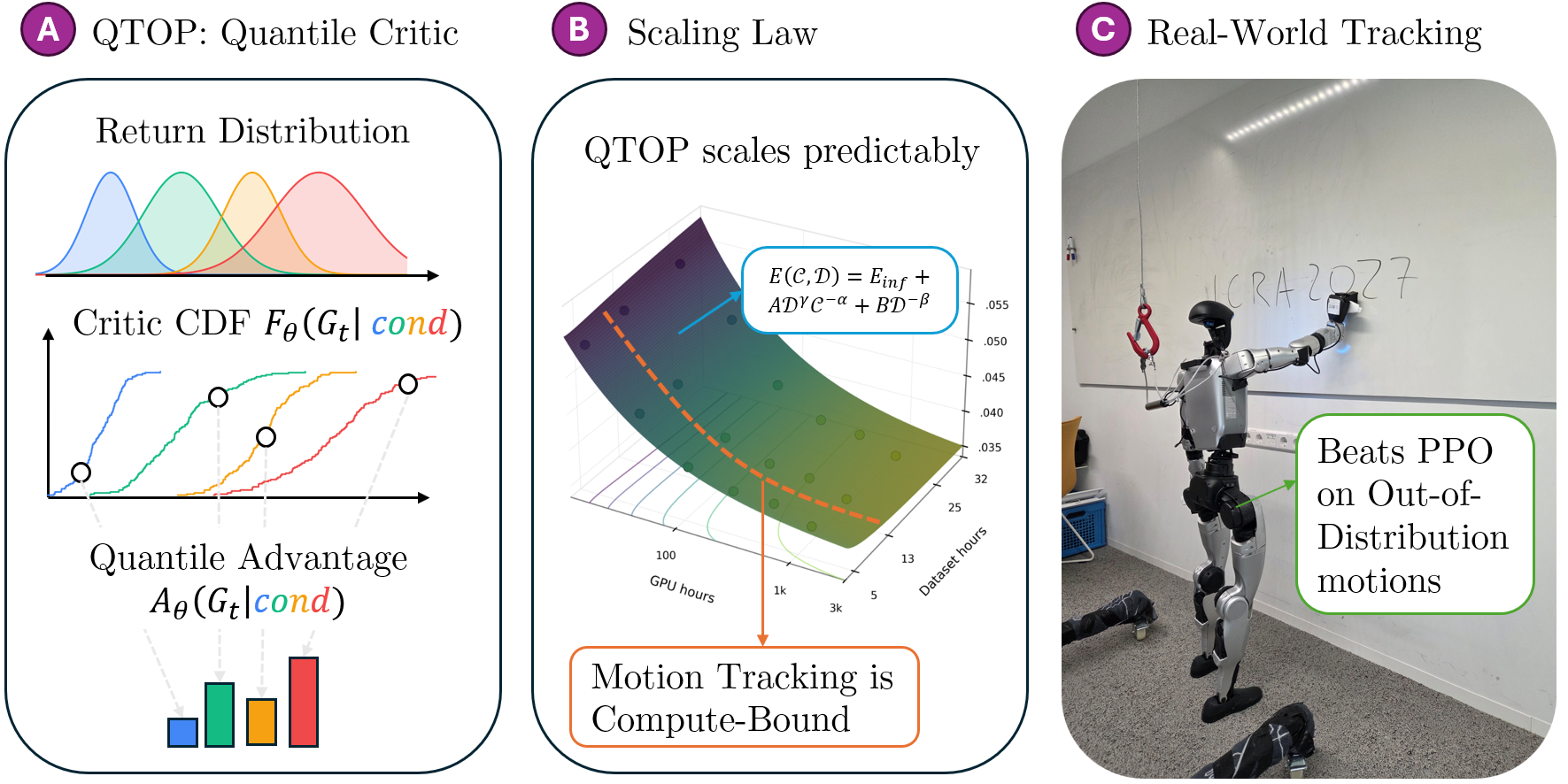}
\captionof{figure}{Achieving high-fidelity real-world tracking with QTOP. We modify PPO to include advantage estimates coming from a distributional critic (A). Using a scaling law in terms of compute (GPU hours) and number of reference motion frames (dataset hours), we analyze the scaling behavior of QTOP and find that motion tracking is heavily compute-bound (B). Finally, we show that QTOP performs well on real-world tracking with minimal data (C).}
\label{fig:teaser}
\end{strip}

\begin{abstract}
Learning human motion tracking policies through reinforcement learning (RL) has emerged as a core capability that underpins a majority of recent progress in humanoid whole-body control. However, achieving generalization to unseen motions at high fidelity requires jointly scaling training to thousands of GPU and dataset hours. This work hypothesizes that motion tracking is inherently compute-bound because training on a diverse set of motions introduces learning interference, similar to multi-task RL. To tackle this, we propose Quantile Tracking Optimization (QTOP), which improves sample efficiency by introducing a distributional advantage estimate conditioned on state and reference pairs. QTOP outperforms on and off-policy RL baselines both in simulation and real-world tracking scenarios on multiple datasets by up to 10\%. Furthermore, we establish a scaling law for QTOP and find that, aside from scaling predictably, dataset size has minimal impact on performance for moderately sized GPU hour regimes.
\end{abstract}
%%%%%%%%%%%%%%%%%%%%%%%%%%%%%%%%%%%%%%%%%%%%%%%%%%%%%%%%%%%%%%%%%%%%%%%%%%%%%%%%
\section{INTRODUCTION}

Humanoid motion tracking aims to learn policies conditioned on reference motions, allowing humanoid robots to generate motor commands that faithfully reproduce diverse behaviors in real time. Over the past few years, motion tracking has rapidly evolved from learning individual skills \cite{liao2025beyondmimic, zhang2025track}  to training generalist policies on thousands of motions  \cite{chen2025gmt, luo2025sonic, yin2026unitracker, qi2026humanoid} with remarkable results on real robot hardware. A key enabler of modern trackers is the utilization of human motion datasets as a rich source of behavioral supervision as well as massively parallelized RL in simulation to converge to a stable and general controller. 

As a result, the central challenge is no longer learning individual motions, but optimizing a single policy over a heterogeneous collection of tracking tasks. While the desired behavior is explicitly specified through reference frames at high frequency, different motions can induce distinct state visitation and return distributions, making optimization challenging \cite{joshi2025benchmarking}. Conventional RL algorithms, such as PPO, typically aggregate these diverse learning signals through a single scalar value function, which can lead to overly conservative policies \cite{wang2026omnixtreme}. The resulting tracking errors, although often small, can accumulate over the course of a rollout and become particularly detrimental for loco-manipulation tasks, where accurate global position, orientation and timing are required to grasp objects with high dexterity. 

A common strategy to eliminate such tracking errors is to scale both motion datasets and simulation throughput \cite{luo2025sonic, qi2026humanoid, zeng2026scaling}. Larger datasets expose policies to a broader range of behaviors, while increased compute enables faster convergence to robust imitation. This scaling recipe has proven highly effective, but it requires substantial compute resources of over \(10^4\) GPU hours, which is out of reach for most robotics researchers. Secondly, it does not address whether additional data and compute contributes equally to performance. We hypothesize in this work that general motion tracking is inherently multi-task, which makes the task compute-bound even at small dataset scales. To address this, we introduce \textbf{Quantile Tracking Optimization} (QTOP), a PPO-based training recipe for large-scale humanoid motion tracking that combines Muon optimization with a distributional value critic. Instead of assigning credit using only a scalar return residual, QTOP estimates the return distribution conditioned on each observation-reference pair and reinforces rollouts according to how favorable their realized returns are relative to this local distribution. This converts tracking difficulty across motions and contact regimes into a calibrated advantage signal, enabling more effective policy improvement.

Using QTOP, we study how motion-tracking performance scales jointly with training compute and motion-dataset size. We construct  nested subsets of the same motion corpus for multiple seeds and train policies over a grid of dataset sizes and optimization budgets, allowing us to separate the marginal benefit of reference data from additional RL compute. Across the low-to-moderate data regime considered in our experiments, we find that increasing the dataset size improves performance only when paired with sufficient compute, while additional training compute remains the dominant bottleneck. Together with existing results at substantially larger data scales \cite{qi2026humanoid}, our findings suggest that an important understudied direction is not only how to collect larger motion corpora, but how to design learning algorithms that can effectively exploit them.

Finally, we evaluate QTOP against strong PPO baselines and representative off-policy alternatives on both academic held-out tracking benchmarks and our own motion capture dataset. QTOP consistently achieves the best tracking performance and transfers to improved real-world tracking on a Unitree G1 under a fixed data and training budget. This points to learning algorithms that can more effectively exploit large reference-conditioned motion datasets as an important direction for future research.

We summarize our contributions as follows:
\begin{itemize}
    \item We identify multitask optimization as a central bottleneck in large-scale humanoid motion tracking.
    \item We introduce QTOP, combining PPO, Muon optimization and a custom distributional critic for advantage estimation using return quantiles for scalable training.
    \item We characterize the joint scaling of tracking performance with dataset size and compute. We show that QTOP continues to benefit from additional compute, whereas increasing dataset size yields comparatively smaller gains in practical regimes.
    \item We find that QTOP consistently outperforms well tuned on- and off-policy RL methods on held-out datasets.
\end{itemize}

\section{RELATED WORK}
\subsection{Motion Tracking}
Learning physical controllers of humanoid characters is a deeply studied problem in computer graphics and robotics. Foundational works quickly evolved from basic RL formulation and single clip learning \cite{peng2018deepmimic, wang2020unicon, wagener2022mocapact}, to larger scale imitation \cite{luo2023perpetual}. This has sparked great interest in improving and deploying trackers for real robot hardware, yielding expressive and dynamic replay across a wide variety of motions and robots \cite{fu2024humanplus, liao2025beyondmimic, xue2025unified, he2025hover, chen2025gmt, ze2025twist, yin2026unitracker}.

Nonetheless, learning a unified tracker that accurately imitates novel, unseen motions remains challenging. Recent works have increasingly turned to scale as a solution. SONIC \cite{luo2025sonic} demonstrates the benefits of scaling up data collection and compute budgets, and Humanoid-GPT \cite{qi2026humanoid} further expands on this philosophy by spending spending 75\% of a $15,000$ GPU hour budget on training $384$ separate PPO experts before distilling it into a single model.
 More closely related to ours, concurrent work Scaling BFM \cite{zeng2026scaling} studies how tracking performance scales with increased GPU budgets and dataset sizes. In contrast, we study the algorithmic foundations of scaling in limited compute regimes of up to $1,300$ GPU hours, introducing QTOP and fitting explicit scaling relationships to disentangle how tracking performance depends jointly on dataset size, training compute, and the learning algorithm.

%The latter studies num GPUs and rollout depth. We need to mention that this supports our findings because deeper rollouts mean less noise in the value function (at the cost of some extra compute). We on the other hand study fixed rollout depth regime and look at the algorithmic side of things.

\subsection{Massively Parallelized RL}
Massively parallel RL utilizes multiple simulation environments to speed up rollout collection alongside a learning policy \cite{nair2015massively,rudin2022learning}. Across most tasks, PPO is the most commonly chosen algorithm due to its stability, scaling properties \cite{hilton2023scaling}, and highly optimized implementations \cite{shengyi2022the37implementation}. While, on the other hand, off-policy methods struggle to scale out of the box \cite{park2026horizon}, there has been a greater effort to revisit off-policy methods for parallel simulated robotics tasks \cite{sabatini2026bridging, seo2025fasttd3, kim2026flashsac, voelcker2026relative} due to its sample efficiency. 

Akin to motion tracking, multitask RL introduces an additional problem of task interference \cite{joshi2025benchmarking}. However, while PPO is the dominant paradigm for motion tracking, most multi-task RL works focus on using off-policy algorithms such as SAC \cite{yu2020meta} and combine it with additional strategies such as gradient surgery  \cite{chen2018gradnorm, liu2023famo} or return normalization \cite{cho2025ars}. TOPPO \cite{li2026toppo} similarly builds on on-policy PPO, but targets discrete multitask RL benchmarks, whereas we address interference within motion tracking, where a continuously changing reference induces a smooth continuum of objectives rather than explicitly separated tasks.

\section{METHOD}
\subsection{Tracking setup}
For the tracker formulation, we largely follow the BeyondMimic recipe \cite{liao2025beyondmimic}. We therefore only define the necessary variables and omit details such as reward design, joint impedance and domain randomization, which have remained unchanged in our work. The goal of motion tracking is to optimize a tracking reward $r_{tracking}$, which consists of position and orientation errors for selected robot bodies with respect to the reference motion over a fixed horizon of $N=500$ steps. The tracking controller runs at $50$ Hz and, at every timestep, receives an observation  
\[
  \mathbf{o}_t =
  \left[
  \mathbf{c}_t,\;
  \boldsymbol{\phi}^{\mathrm{ref}}_t,\;
  \boldsymbol{\omega}^{\mathrm{base}}_t,\;
  \mathbf{q}_t - \bar{\mathbf{q}},\;
  \dot{\mathbf{q}}_t,\;
  \mathbf{a}_{t-1}
  \right]. \]
Here, \(\mathbf{c}_t=[\mathbf{q}^{\mathrm{ref}}_t,\dot{\mathbf{q}}^{\mathrm{ref}}_t]\)
is the reference joint position and velocity command, consisting of retargeted motion sequences which we obtain via GMR \cite{araujo2025retargeting}. Furthermore, \(\boldsymbol{\phi}^{\mathrm{ref}}_t \in \mathbb{R}^6\) is a 6D
orientation error between the robot torso and reference torso, represented by
the first two columns of the relative rotation matrix, while the remaining terms are
proprioceptive: base angular velocity \(\boldsymbol{\omega}^{\mathrm{base}}_t\), joint positions relative to the nominal
pose \(\mathbf{q}_t - \bar{\mathbf{q}}\), joint velocities \(\dot{\mathbf{q}}_t\), and the previous action \(\mathbf{a}_{t-1}\). We omit including the base robot position and linear velocities in \(\mathbf{o}_t\) for consistency between simulation and real-world experiments, as in the latter case an external state estimation module would be required. Our embodiment of choice is the 29-DoF Unitree G1 humanoid with Dex3-1 hands, which yields \(\mathbf{o}_t \in \mathbb{R}^{154}\) as finger joints are not modeled explicitly.

\subsection{Quantile Advantage Function}
Motion tracking can be formulated as a reference-conditioned Markov Decision Process. At the start of each training episode we sample a reference motion segment \(\mathbf{c}_{1:N} \sim p_{\mathrm{data}}\). The policy \(\pi_\theta\) is conditioned on the observation \(\mathbf{o}_t\) and aims to maximize
\[
J(\theta)
=
\mathbb{E}_{\mathbf{c}_{1:N}\sim p_{\mathrm{data}}}
\mathbb{E}_{\pi_\theta}
\left[
\sum_{t=0}^{N-1}
\gamma^t r_{\mathrm{track}}(s_t,\mathbf{c}_t)
\right],
\]
where \(s_t\) is the simulator state and \(\gamma\in[0,1)\) is the discount factor. The expectation over reference motions makes large-scale tracking a continuous
multi-task optimization problem. Unlike many multi-task RL settings, where each
task is specified by a discrete identifier, the task in motion tracking is indirectly specified
by the reference command sequence \(\mathbf{c}_{1:N}\). Moreover, the conditioning changes at every timestep through \(\mathbf{c}_t\), while the actor observes only a compact local command rather than an explicit motion identity and future trajectory. Thus, similar observations can correspond to different future command
continuations, contact regimes, and return targets. This results in substantially different state visitation and return distributions across samples of \(\mathbf{c}_{1:N}\), while sharing the same reward function and policy parameters. Prior work has aimed to address this issue by adding future reference frames into the observation at the cost of latency \cite{luo2025sonic}, or decomposing training across
motion subsets or expert policies and subsequently distilling them into a
single deployable policy~\cite{liao2025beyondmimic,wang2026omnixtreme,qi2026humanoid}.

\begin{figure}
    \vspace{5pt}
    \centering
    \includegraphics[width=1.0\linewidth]{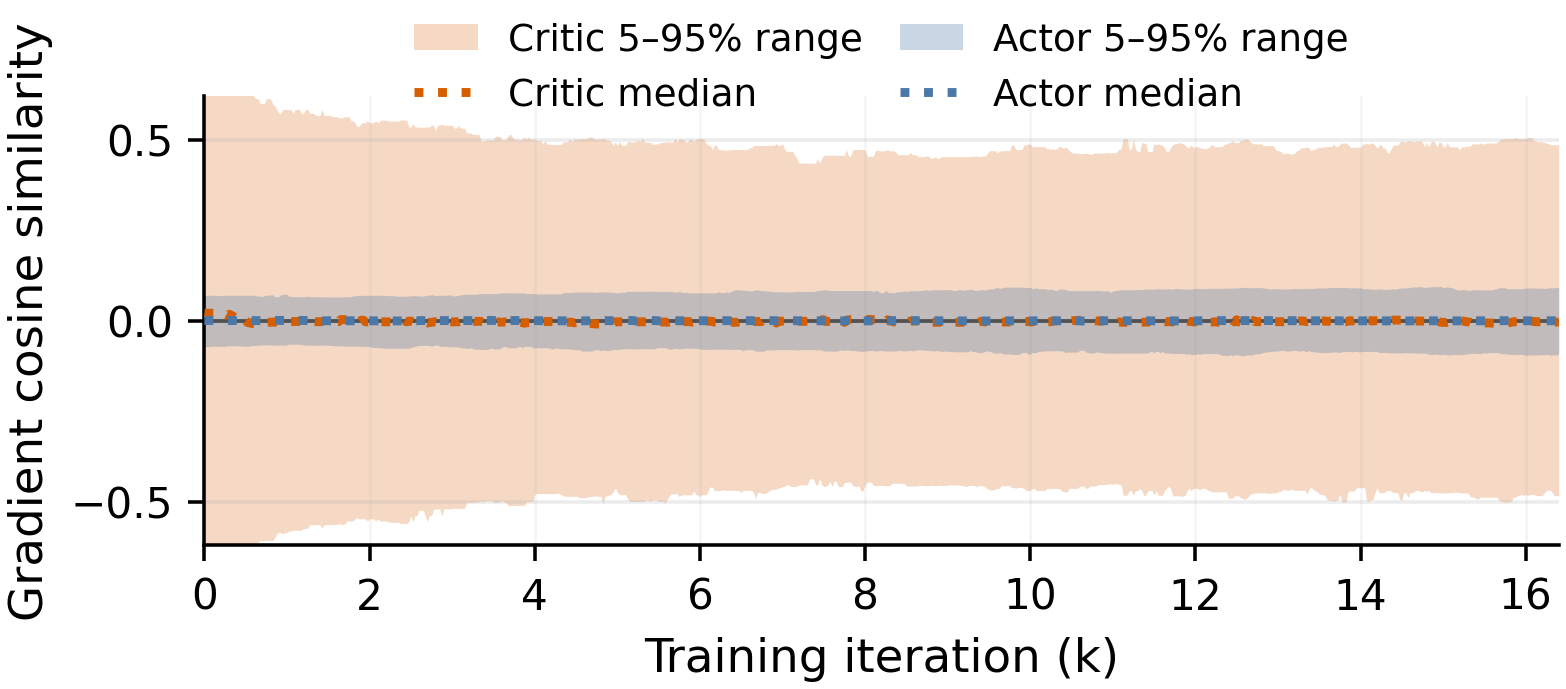}
  \caption{Per-motion actor and critic gradient cosine similarities during
  training. At each point, gradients are computed for individual
  reference motions at a common parameter snapshot. Shaded bands denote the
  5th--95th percentile over motion pairs, and dotted lines denote the median.
  Critic gradients exhibit substantially greater variation,
  including a stronger negative tail, indicating more severe disagreement in value learning.}
    \label{fig:grad_conflict}
\vspace{-15pt}
\end{figure}

In this work, on the other hand, we aim to study a more robust learning algorithm that can directly alleviate optimization issues when sampling over large command spaces. We build on top of PPO due to its success in motion tracking and existing highly optimized baselines, which optimizes $J(\theta)$ using rollout samples and an advantage estimate \(\hat{A}_t\).
Typically, the policy update is driven by a clipped surrogate objective whose learning signal is
proportional to \(\hat{A}_t\), which is computed from a learned critic \(V_\psi\), e.g., through Generalized Advantage Estimation (GAE) \cite{schulman2015high}.
%\[
%\hat{A}^{\mathrm{GAE}}_t
%=
%\sum_{l=0}^{N-t-1}
%(\gamma\lambda)^l
%\delta_{t+l}, \ \ \ \ 
%\delta_t =
%r_t + \gamma V_\psi(\mathbf{o}_{t+1}) - V_\psi(\mathbf{o}_t).
%\]
Inspired by previous work \cite{joshi2025benchmarking, li2026toppo}, which indicates that critic learning is a key bottleneck in actor-critic multi-task settings, we first analyze cosine similarities of actor and critic gradients respectively, and find that the same holds for motion tracking in Figure~\ref{fig:grad_conflict}. This encourages rethinking how the critic is used for policy updates. Standard GAE compresses the critic into a scalar mean value and forms advantages from scalar temporal difference residuals. When value learning is difficult, these residuals can be detrimental for updating the actor in the right direction. 

Rather than attempting to align these critic gradients directly, QTOP changes how critic predictions are converted into policy learning signals.
QTOP learns a distributional critic and computes a distributional advantage using the relative position of returns with respect to the predicted return distribution. This reduces sensitivity to motion-dependent return scales and critic errors, resulting in more balanced contributions to the aggregate actor update. Formally, QTOP replaces the scalar critic \(V_\psi(\mathbf{o}_t)\) with a
categorical distributional critic over a fixed return support \cite{bellemare2017distributional}
\[
\mathcal{Z}=\{z_1,\ldots,z_K\}, \qquad z_1 < \cdots < z_K.
\]
Given an observation \(\mathbf{o}_t\), the critic outputs logits
\(\ell_\psi(\mathbf{o}_t)\in\mathbb{R}^K\), which define probabilities
\[
p_\psi^i(\mathbf{o}_t)
=
\frac{\exp(\ell_\psi^i(\mathbf{o}_t))}
{\sum_{j=1}^K \exp(\ell_\psi^j(\mathbf{o}_t))}.
\]
The predicted return distribution is
\[
Z_\psi(\mathbf{o}_t)
=
\sum_{i=1}^K p_\psi^i(\mathbf{o}_t)\,\delta_{z_i},
\]
and its expectation recovers a scalar value estimate
\[
V_\psi(\mathbf{o}_t)
=
\mathbb{E}[Z_\psi(\mathbf{o}_t)]
=
\sum_{i=1}^K p_\psi^i(\mathbf{o}_t) z_i.
\]
We use this expected value wherever PPO requires a scalar critic, e.g., for
bootstrapping returns in non-terminal rollouts.

To compute the policy advantage, QTOP evaluates where the realized return lies
under the predicted local return distribution. Let
\[
F_\psi(x\mid \mathbf{o}_t)
=
\sum_{i:z_i\leq x} p_\psi^i(\mathbf{o}_t)
\]
denote the CDF of the predicted return distribution, with linear interpolation
between adjacent support atoms. For each rollout sample with (bootstrapped) returns \(\hat{G}_l\), we compute
\[
u_t = F_\psi(\hat{G}_t \mid \mathbf{o}_t),
\]
and define the distributional advantage
\[
\hat{A}^{\mathrm{dist}}_t
=
2\left(u_t-\frac{1}{2}\right).
\]
Thus, \(\hat{A}^{\mathrm{dist}}_t>0\) when the realized return is above the
median of the critic's predicted local return distribution, and
\(\hat{A}^{\mathrm{dist}}_t<0\) when it is below. In practice, we use a warmup phase with standard GAE advantages and then blend the standardized scalar and distributional advantages:
\[
\hat{A}^{\mathrm{QTOP}}_t
=
(1-\alpha)\,\mathrm{Norm}\!\left(\hat{A}^{\mathrm{GAE}}_t\right)
+
\alpha\,\mathrm{Norm}\!\left(\hat{A}^{\mathrm{dist}}_t\right),
\]
where \(\alpha\in[0,1]\) controls the contribution of the quantile advantage.
The resulting advantage is normalized over the rollout batch and used in the
standard PPO surrogate objective, which is unchanged. Unless otherwise stated, we use \(\alpha=0.5\) and enable the quantile advantage after a 500-iteration warmup. Finally, we combine this distributional advantage with the Muon optimizer \cite{jordan2024muon}, as we empirically find it to provide more efficient optimization than Adam when combined with distributional critics.

\section{EXPERIMENTS}
We evaluate QTOP along three dimensions. Firstly, we compare tracking performance in simulation against strong on-policy and off-policy RL baselines under a fixed compute budget. Secondly, we establish a scaling law with respect to motion dataset and RL compute hours. Finally, we deploy the resulting policies on a Unitree G1 humanoid and evaluate real-world tracking behavior. For our training dataset, we use the union of AMASS \cite{AMASS:ICCV:2019} and LAFAN1 \cite{harvey2020robust}. Although this corpus is significantly smaller than those used by recent large-scale systems \cite{luo2025sonic, qi2026humanoid}, our objective is not to build a state-of-the art foundation model, but to isolate the effect of the learning algorithm under limited data and compute budgets. As shown in Section~\ref{sec:scaling}, even within this comparatively small data regime, tracking performance requires substantial compute to approach saturation.

%We train on a dataset comprising of AMASS \cite{AMASS:ICCV:2019} and LAFAN1 \cite{harvey2020robust}. We acknowledge that this is significantly smaller than recent works \cite{luo2025sonic, qi2026humanoid} we don't want to build the best tracker possible, but rather highlight the strength of QTOP as an algorithm. Furthermore, in section \ref{sec:scaling} we show that despite this small dataset sizes, there is a lot of compute needed to saturate performance even with QTOP. In the following we aim show three points. QTOP beats SOTA RL algorithms. We can establish a scaling law and find a relationship between data needed versus compute and finally, QTOP performs well on real robots. 

\begin{table}[t]
\vspace{5pt}
  \centering
  \scriptsize
  \setlength{\tabcolsep}{6.2pt}
    \renewcommand{\arraystretch}{1.05}
  \newcommand{\sem}[1]{\ensuremath{{\scriptstyle \pm #1}}}
  \caption{Algorithm comparison on two datasets. Position errors are
  reported in millimeters or centimeters and orientation errors in degrees.
  Lower is better for all metrics.}
  \label{tab:tracking_eval_metrics}
  \begin{tabular}{@{}lcccc@{}}
    \toprule
    Method
    & \shortstack{MPKPE\\(mm) $\downarrow$}
    & \shortstack{R-MPKPE\\(mm) $\downarrow$}
    & \shortstack{Global EE Pos.\\(cm) $\downarrow$}
    & \shortstack{EE Ori.\\(degrees) $\downarrow$} \\
    \midrule
    \multicolumn{5}{l}{\textbf{MOSAIC (optical mocap, 3.1h)}} \\
      QTOP ($\alpha{=}0.5$)
        & \textbf{41.84}\sem{0.90}
        & \textbf{38.12}\sem{0.90}
        & 36.66\sem{1.82}
        & \textbf{11.44}\sem{0.00} \\
      QTOP ($\alpha{=}0$)
        & 43.17\sem{0.22}
        & 39.53\sem{0.27}
        & \textbf{32.71}\sem{0.87}
        & 13.37\sem{1.44} \\
      PPO (Muon)
        & 44.99\sem{0.14}
        & 41.70\sem{0.08}
        & 35.75\sem{0.87}
        & 11.86\sem{0.49} \\
      PPO (Adam)
        & 45.68\sem{0.56}
        & 41.96\sem{0.55}
        & 41.88\sem{0.19}
        & 13.12\sem{0.11} \\
      SAC
        & 54.56\sem{0.89}
        & 51.48\sem{0.73}
        & 77.90\sem{8.47}
        & 14.64\sem{0.39} \\
      FastTD3
        & 68.79\sem{0.97}
        & 64.32\sem{0.74}
        & 50.33\sem{2.14}
        & 20.54\sem{0.06} \\
    \addlinespace[0.35em]
    \midrule
    \addlinespace[0.35em]
    \multicolumn{5}{l}{\textbf{Ours (inertial mocap, 1h)}} \\
      QTOP ($\alpha{=}0.5$)
        & \textbf{39.76}\sem{1.58}
        & \textbf{37.70}\sem{1.55}
        & 45.33\sem{2.36}
        & \textbf{13.10}\sem{0.10} \\
      QTOP ($\alpha{=}0$)
        & 41.79\sem{0.25}
        & 40.07\sem{0.62}
        & \textbf{39.83}\sem{3.05}
        & 14.32\sem{0.97} \\
      PPO (Muon)
        & 44.65\sem{0.46}
        & 42.20\sem{0.00}
        & 41.47\sem{1.51}
        & 14.08\sem{1.20} \\
      PPO (Adam)
        & 45.72\sem{0.10}
        & 43.59\sem{0.41}
        & 45.61\sem{0.08}
        & 14.78\sem{0.19} \\
      SAC
        & 56.42\sem{2.00}
        & 55.30\sem{2.47}
        & 86.74\sem{0.88}
        & 15.97\sem{0.44} \\
      FastTD3
        & 73.55\sem{0.49}
        & 70.76\sem{0.10}
        & 65.01\sem{3.84}
        & 24.52\sem{0.38} \\
    \bottomrule
  \end{tabular}
  \vspace{-12pt}
\end{table}

\subsection{Simulation Evaluation}
\textbf{Baselines and training budget.} 
We train QTOP and several baselines for a fixed compute budget of 200 GPU hours, corresponding to around 24 hours of parallel training on a single node of eight A100 GPUs. Wherever possible, we share the exact network size and architecture. We follow previous work \cite{luo2025sonic} to choose an appropriate number of learnable parameters and implement an MLP with 4 hidden layers, equating to $11.5$ million parameters in total. Furthermore, we use mjlab \cite{zakka2026mjlablightweightframeworkgpuaccelerated} as our simulator due to its open-source availability and throughput advantages. 

Our on-policy comparisons isolate the contributions of the optimizer, distributional critic, and quantile advantage. \emph{PPO (Adam)} is the standard motion tracking baseline, whereas \emph{PPO (Muon)} replaces the optimizer with Muon \cite{jordan2024muon} while retaining the scalar critic. \emph{QTOP ($\alpha=0$)} uses Muon and the categorical distributional critic, but computes GAE using the expected value of the predicted return distribution. Finally, \emph{QTOP ($\alpha=0.5$)} adds the distributional advantage with equal weight. Additionally, we compare against \emph{SAC} using the recently released optimized RSL-RL implementation~\cite{sabatini2026bridging} and \emph{FastTD3}~\cite{seo2025fasttd3} due to its focus on humanoid tasks and inclusion of a distributional critic. We also experimented with MT-GRPO~\cite{joshi2025benchmarking}, which was motivated by the impaired critic gradients of Figure~\ref{fig:grad_conflict}, but found its reliance on complete Monte Carlo rollouts prohibitively expensive for our 10 second (500-step) tracking horizon.   

\begin{figure*}[t]
    \vspace{5pt}
    \centering
    \includegraphics[width=1.0\linewidth]{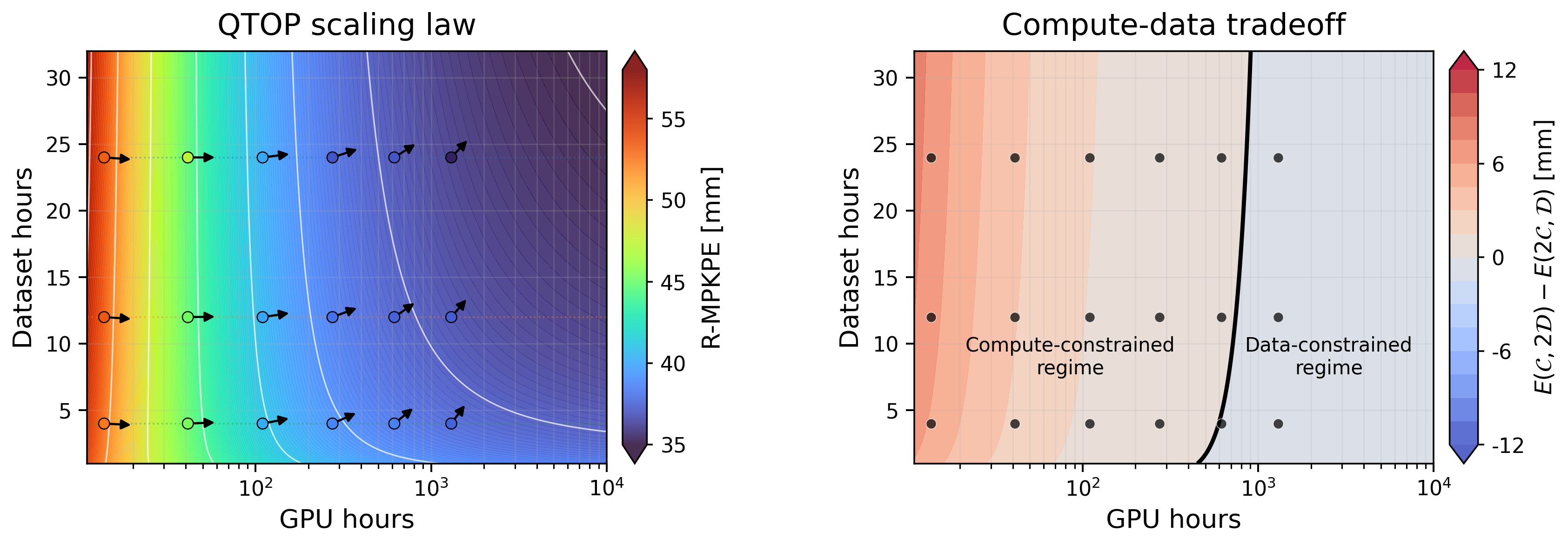}
    \caption{Left: Fitted scaling law surface to log-spaced compute points for three different dataset sizes. Right: R-MPKPE difference heatmap of doubling compute versus doubling dataset hour budgets. For reasonable compute budgets up to 700 GPU hours, dataset size has a marginal impact on performance.}
    \label{fig:scaling_law_2d}
    \vspace{-12pt}
\end{figure*}

\textbf{Evaluation Datasets.} We evaluate generalization performance on unseen motion sequences. We use the 3.1-hour optical motion-capture subset of MOSAIC \cite{sun2026mosaic}, whose motions produce consistently higher-quality retargeting than MOSAIC's inertial motion capture subset. To broaden evaluation beyond this, we additionally collect our own inertial motion-capture dataset of diverse human motions including everyday activities such as walking, crouching, jumping, cooking, shelf restocking and sports. For each policy, we evaluate randomly sampled reference motions using $8192$ parallel environments and report the mean and standard error across two independently trained seeds.

\textbf{Results.} Table~\ref{tab:tracking_eval_metrics} reports mean Per-Keybody Position Error (MPKPE), root relative MPKPE (R-MPKPE), global end-effector position (Global EE Pos.) as well as global end-effector orientation (EE Ori.) errors. Following BeyondMimic, we define key bodies as 14 representative rigid links
  spanning the pelvis, torso, arms, and legs, rather than all kinematic joints, such that MPKPE measures anchor-aligned position error over these key bodies and R-MPKPE measures body positions relative to the torso. The end effector metrics isolate the robot's hands and feet positions, which are particularly relevant for loco-manipulation. QTOP ($\alpha=0.5$) proves to have the strongest overall generalization performance, achieving the lowest MPKPE, R-MPKPE and end-effector orientation error, whereas QTOP ($\alpha=0$) obtains the lowest global end-effector position error. We note that the global end-effector position errors accumulate over time and therefore are comparatively large, since they are measured in the world frame and the lack of global root position as policy inputs. This indicates that while the distributional critic is useful in itself, utilizing a mix of GAE and distributional advantage estimation can further improve tracking performance overall, with QTOP ($\alpha=0.5$) consistently yielding about $10\%$ improvement over the standard PPO baseline.

Interestingly, we find off-policy algorithms to perform underwhelmingly in motion tracking despite significant tuning efforts. While SAC worked well out of the box with similar hyperparameters as PPO, its performance saturates well before that of PPO. For FastTD3, on the other hand, we employed an automated agent based on GPT 5.5 with a budget of 100  training runs to derive both optimized hyperparameters as well as having the option to change the implementation. The agent's starting point was the original FastTD3 codebase as well as our distributional critic support hyperparameters from QTOP. Despite these significant tuning efforts, its performance lags behind drastically, suggesting that motion tracking poses additional challenges beyond those introduced by learning high DoF humanoid controllers.

\subsection{Scaling Law}
\label{sec:scaling}
\textbf{Experiment Setup.}
 The previous experiments evaluate QTOP under a fixed compute budget. However, an effective learning algorithm should not only perform well at a single operating point, but should also continue to improve as additional resources become available. Inspired by the vast literature on scaling laws for language modeling \cite{hoffmann2022training} and, more recently, reinforcement learning \cite{hilton2023scaling, fu2026compute}, we study how the tracking performance of QTOP scales jointly with respect to training compute and motion data. Specifically, we fit a predicted error surface as 

\[E(\mathcal{C},\mathcal{D}) = E_0 + A\mathcal{D}^{\gamma}\mathcal{C}^{-\alpha} + B\mathcal{D}^{-\beta},\]
where ($\mathcal{C}$) denotes GPU-hours and ($\mathcal{D}$) denotes dataset hours. The constant \(E_0\) represents an irreducible error floor. Traditionally, scaling laws do not feature the interaction component $\mathcal{D}^\gamma$, but we include it due to providing a better fit that better captures optimization error. Overall,   \(A\mathcal{D}^{\gamma}\mathcal{C}^{-\alpha}\) always decreases with compute, but increases with dataset size as a larger motion corpus can require more optimization, while $B\mathcal{D}^{-\beta}$ captures the typical behavior of data scaling.

We focus on compute and dataset size while holding the model architecture fixed, allowing us to isolate which resource provides the larger marginal improvement. Prior motion-tracking work has generally increased data and compute simultaneously, making their individual contributions difficult to disentangle. SONIC~\cite{luo2025sonic}, for example, reports separate curves for increasing dataset size and increasing compute, but does not evaluate an interaction over a complete compute and data grid. Our study instead varies both quantities systematically, enabling us to characterize their joint effect and compare the benefits of allocating additional resources to either data or optimization.

\textbf{Training and evaluation split.} We construct three training sets containing approximately \(4\), \(12\), and \(24\) hours of motion data, where the \(24\)-hour set corresponds to the full training corpus used in the previous experiments. For the \(4\)- and \(12\)-hour settings, we independently sample three subsets of the corresponding size from the full corpus to eliminate sampling bias. We train QTOP three times for different seeds at each dataset size and evaluate checkpoints at log-spaced compute intervals. At each checkpoint, we then measure R-MPKPE on the held-out MOSAIC test set and average the metric across the three runs. This produces an evaluation grid indexed by dataset size and accumulated GPU-hours.

To reduce sensitivity to noisy checkpoint evaluations, we estimate the scaling-law parameters using a robust soft-\(\ell_1\) loss \cite{hoffmann2022training}. We then visualize the resulting surface using iso-error contours and local compute-data improvement directions and additionally perform leave-one-dataset-size-out validation to assess whether the fitted relationship generalizes across dataset scales rather than merely interpolating individual measurements.

\begin{figure}[t]
    \centering
    \includegraphics[width=1.0\linewidth]{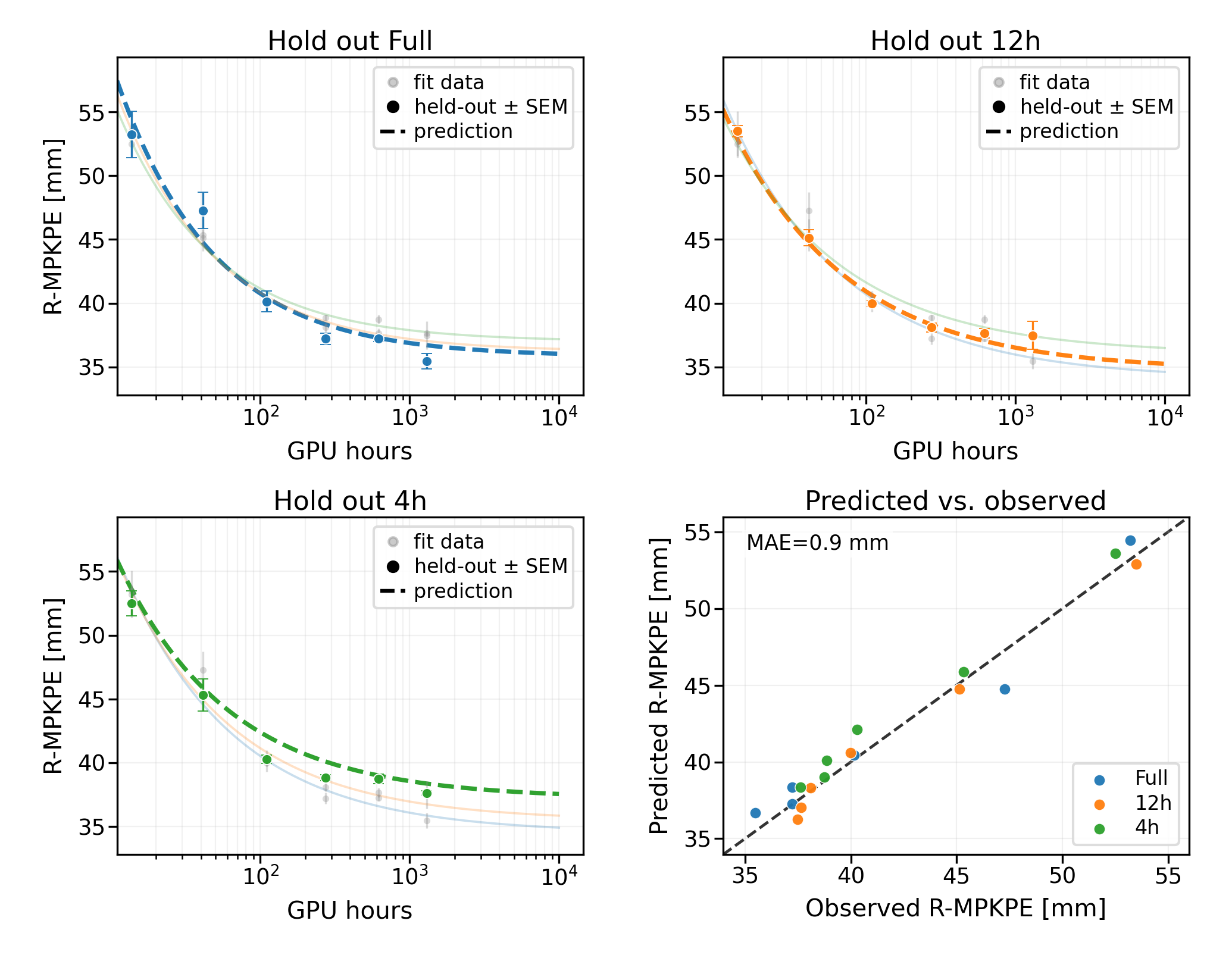}
    \caption{Extrapolating scaling law predictions to each dataset size yields reliable predictions of tracking error. Bottom right: Predicted versus observed R-MPKPE for each held-out dataset size when fit to the remaining two.}
    \label{fig:scaling_law_generalization}
    \vspace{-12pt}
\end{figure}

\textbf{Interpolation Results.} Figure~\ref{fig:scaling_law_2d} visualizes the fitted compute-data scaling law for QTOP. The robust fit achieves \(R^2=0.981\) on the log-spaced evaluation grid, with fitted exponents \(\alpha =0.592, \beta= 0.0277, \gamma = 0.0915.\)
We use the fitted surface \(E(\mathcal{C},\mathcal{D})\) to compare two local resource-allocation strategies: doubling the dataset size, yielding \(E(\mathcal{C},2\mathcal{D})\), or doubling the training compute \(E(2\mathcal{C},\mathcal{D})\). The right panel plots the difference \(E(\mathcal{C}, 2\mathcal{D})-E(2\mathcal{C},\mathcal{D}),\) such that positive values indicate that doubling compute is predicted to produce lower tracking error than doubling dataset hours. The left panel shows iso-error contours of the same fitted surface, with arrows indicating the locally preferred resource-allocation direction. 

Across most of the evaluated low-to-moderate compute regime, extending to approximately \(700\) GPU-hours, the fitted law predicts that increasing training compute is more beneficial than increasing dataset size. Thus, even when using QTOP on a comparatively small motion corpus, policy learning remains strongly compute-constrained. These results suggest that simply collecting more reference motions may provide limited benefits unless accompanied by a sufficient compute budget, and motivate further research into RL algorithms that exploit larger datasets more compute-efficiently.

\textbf{Extrapolation Results.} Figure~\ref{fig:scaling_law_generalization} analyzes the generalization capabilities by fitting our law to a subset of the available points. Taking the same training and evaluation setup, we now omit one of the three dataset sizes to fit the parameters $\alpha, \beta, \gamma, E_0$ on the remaining 12 points and evaluate on a respective held-out test set of 6 points. For each dataset, the dashed line represents predicted R-MPKPE behavior as GPU hours increase, whereas solid points indicate empirical observations. Overall, we see that despite the small training set, our law is able to consistently predict the behavior for different dataset sizes with an overall MAE of $0.9$mm.

We want to note that this accuracy should be interpreted within the measured regime: dataset size spans only \(4\)--\(24\) hours, whereas compute varies over nearly two orders of magnitude, and the
fitted surface is comparatively flat along the data axis. We therefore do not claim that the law extrapolates to substantially larger motion corpora. Nevertheless,
Figure~\ref{fig:scaling_law_2d} indicates that even these small datasets require considerable compute before additional data becomes beneficial.
Table~\ref{tab:compute_data_intervention_gains} quantifies this imbalance and compares it with the trends reported by SONIC~\cite{luo2025sonic}. We compute the average gain per doubling by dividing the (unrounded) relative endpoint error reduction by \(\log_2(B_{\max}/B_{\min})\), where $B$ denotes the compute or data budget. Compute scaling provides larger
gains in both settings, while SONIC observes particularly small improvements from substantially increasing its dataset. Our larger data-scaling gain is consistent with operating
in a smaller-data regime. Because SONIC evaluates its data sweep at a fixed maximum compute budget and uses a different error metric, we treat this comparison as qualitative
evidence rather than a controlled benchmark.

  \begin{table}[t]          
  \vspace{5pt}
    \centering
    \scriptsize
    \setlength{\tabcolsep}{2.6pt}
    \caption{
Relative error reduction per budget doubling when scaling compute or data. Frames are converted to hours at 50 Hz.
    }
    \label{tab:compute_data_intervention_gains}
    \begin{tabular}{llcccc}
      \toprule
      System & Intervention & $\mathcal{C}$ (GPUh) & $\mathcal{D}$ (hours)
      & Error & Gain/doubling \\
      \midrule
      Ours
      & Compute
      & 13.7--1303
      & 25
      & .055$\to$.035
      & 5.59\% \\
      Ours
      & Data
      & 1303
      & 5--25
      & .037$\to$.035
      & 3.31\% \\
      SONIC
      & Compute
      & 2k--21k
      & 556
      & .027$\to$.024
      & 3.10\% \\
      SONIC
      & Data
      & 21k
      & 22--556
      & .024$\to$.024
      & 0.53\% \\
      \bottomrule
    \end{tabular}
    \vspace{-12pt}
  \end{table}

\subsection{Motion Contribution Balancing}

%\begin{figure}
%    \centering
%    \includegraphics[width=1.0\linewidth]{figures/distadv_doubling_compute_vs_data_tradeoff_log_subset_robust.png}
%    \caption{Caption}
%    \label{fig:compute_vs_data}
%\end{figure}

%\begin{figure}
%    \centering
%    \includegraphics[width=1.0\linewidth]{figures/distadv_scaling_validation_leave_one_dataset_out_log_subset_robust.png}
%    \caption{Caption}
%    \label{fig:leave_one_out_dataset_scaling}
%\end{figure}

\begin{figure}[b]
    \vspace{-10pt}
    \centering
    \includegraphics[width=1.0\linewidth]{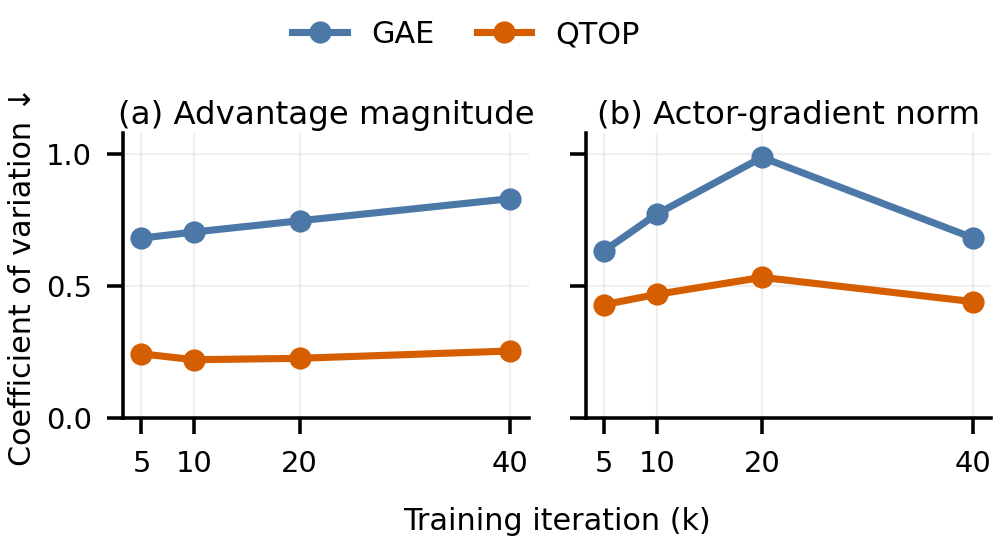}
    \caption{QTOP balances policy updates across motions. Using identical rollouts and frozen policy checkpoints, we compare the across-motion coefficient of variation of advantage magnitudes and actor-gradient norms.}
    \label{fig:qtop_motion_balance}
\end{figure}
To understand why QTOP improves tracking, we compare GAE and QTOP on identical
rollouts from 32 motions while keeping the policy fixed. Figure~\ref{fig:qtop_motion_balance}
shows how unevenly the resulting learning signals are distributed across
motions. QTOP reduces the variation in advantage magnitude by \(68\%\) and in
actor-gradient norm by \(39\%\), averaged across checkpoints. Above, we used gradient cosine similarity as evidence of heterogeneous value learning,
rather than as a quantity that QTOP directly optimizes, as QTOP does not explicitly align motion-conditioned gradients. Instead, it limits how strongly differences in return scale and critic error affect their contribution to the actor update.
This mechanism is complementary to gradient-surgery methods \cite{li2026toppo}, which directly
modify gradient directions but require additional per-task gradient computation. Our method, on the other hand, retains the scalability and minibatch structure of PPO.

\begin{figure}[t]
    \vspace{5pt}
    \centering
    \includegraphics[width=1.0\linewidth]{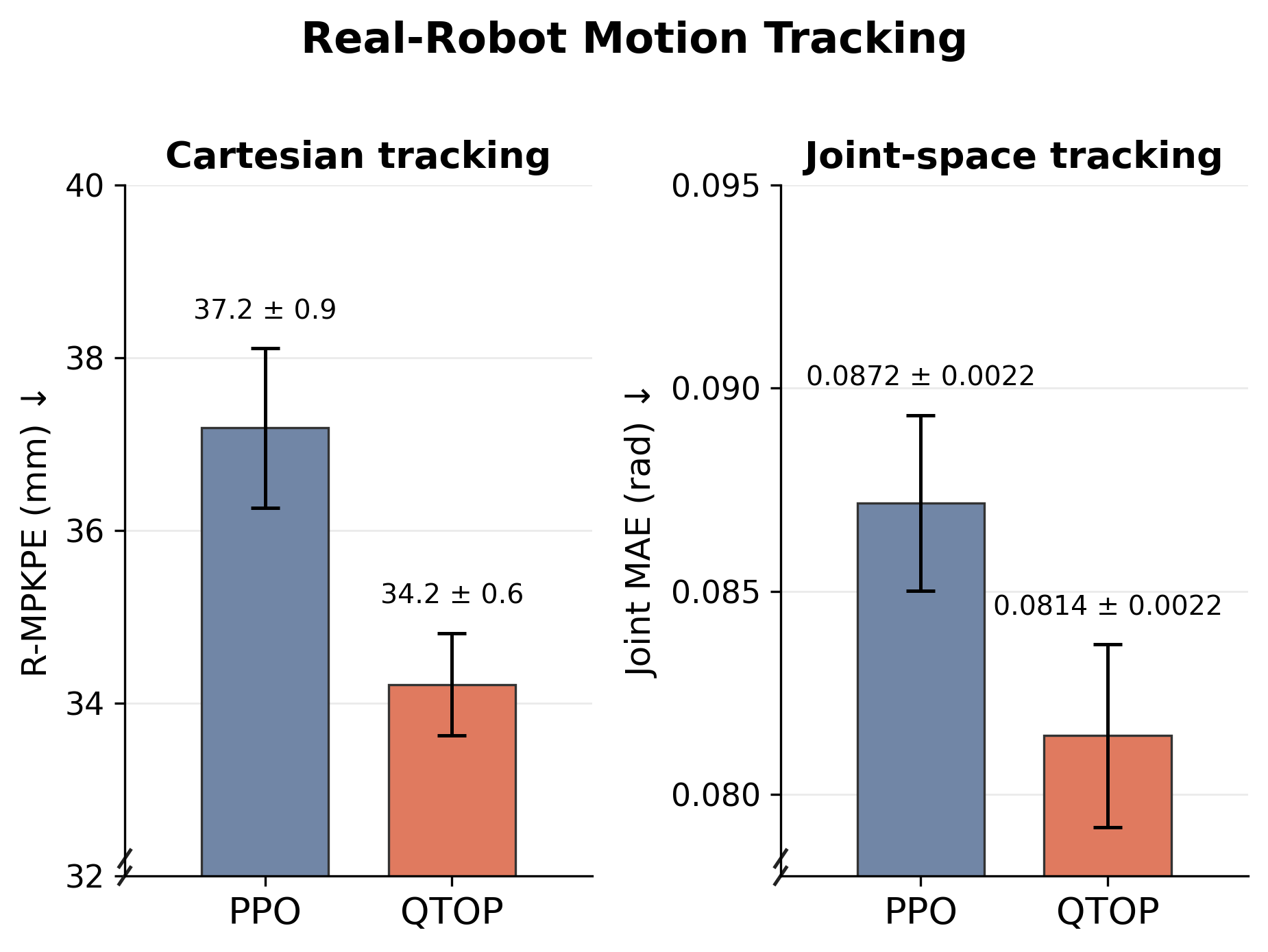}
    \caption{Real world tracking results for converged PPO and QTOP checkpoints in terms of R-MPKPE and joint MAE. QTOP outperforms PPO at the same compute and dataset budgets in unseen motion replay.}
    \label{fig:realworld_results}
    \vspace{-12pt}
\end{figure}

\subsection{Real World Evaluation}
Finally, we investigate whether QTOP’s simulation gains translate to improved tracking on physical hardware. To do so, we deploy our PPO and QTOP policies on a Unitree G1 humanoid, with policy inference running at \(50\mathrm{Hz}\) on an external workstation. We evaluate on 32 diverse motions from our
inertial motion-capture dataset, ranging from 5 to $15\mathrm{s}$ in duration. The selected motions form a representative and diverse subset including walking in different styles, shelf restocking, whiteboard cleaning, playing pool and cooking. For each trial, the robot is initialized at the first reference frame and tracks the motion until completion while its achieved pose is recorded. Each motion is repeated three times for both
methods using checkpoints trained on our 24-hour motion dataset with $10^3$ GPU-hours of compute.

Figure~\ref{fig:realworld_results} summarizes the resulting real-world tracking accuracy in terms of R-MPKPE and absolute joint error (joint MAE), averaged across trials. We note that joint MAE is computed over all actuated
joints, whereas R-MPKPE is evaluated only on the key bodies included in the training reward. We find that QTOP significantly improves both metrics over PPO, reducing R-MPKPE by approximately (8\%), with a smaller but consistent improvement in joint MAE. These results indicate that the tracking improvements obtained by QTOP in simulation transfer to physical hardware under the same data and training-compute budget. 

\begin{figure}[t]
    \vspace{3pt}
    \centering
    \includegraphics[width=1.0\linewidth]{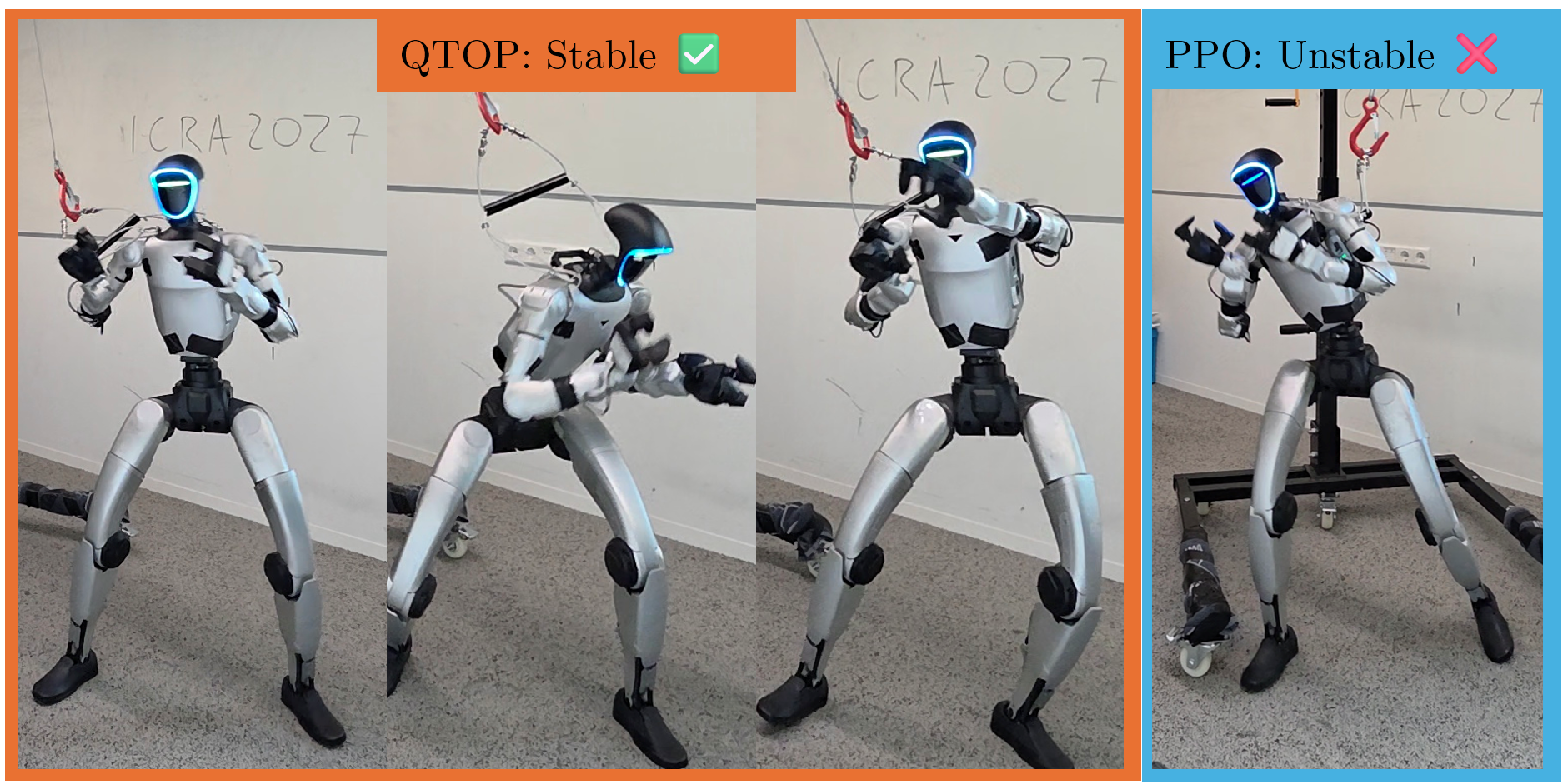}
    \caption{QTOP faithfully replays challenging out-of-distribution motions such as martial arts, whereas PPO relies on multiple stabilization steps.}
    \label{fig:realworld_placeholder}
    \vspace{-10pt}
\end{figure}

Finally, we qualitatively evaluate both policies by replaying challenging out-of-distribution motions, such as martial arts sequences in Figure~\ref{fig:realworld_placeholder}. While both policies produce robust behavior overall, we find that the PPO policy tends to rely on additional stabilization steps during high-velocity upper body movements, leading to a substantial positional drift over time. QTOP, on the other hand, follows the commanded motion more faithfully. We also test a whiteboard cleaning task as shown in Figure~\ref{fig:teaser}. In this scenario, the differences are more visually subtle, but we find that the marginal tracking gains by QTOP can significantly improve the positioning of the robot end effectors. As a result, we conclude that even modest improvements in tracking accuracy can translate into meaningful gains for real-world loco-manipulation tasks.

%Since the policy does not receive any information about future reference frames, it is made to run in real time with minimal latency. Qualitatively speaking, even though both PPO and QTOP yield robust behaviors, we notice that certain out-of-distribution steps or walking are tricky for the motion tracker to follow. However, we find that QTOP generally is better at instruction following, whereas PPO has a more conservative behavior and tends to miss steps or stumble an extra step. Visually the difference is small but for loco-manipulation this can make a big difference.

%We want to see if QTOP can achieve the same gains in the real world that we observed in Table~\ref{tab:tracking_eval_metrics}. We test on the Unitree G1 robot, running the policy on a physical workstation at 50hz. For this test, we take motions from our own inertial mocap dataset and replay them from start to finish. The robot is initialized to the first frame of the motion and subsequently plays until the last frame is reached. During this time, we collect the achieved robot pose which can be then compared against the desired reference. 

%Figure~\ref{fig:realworld_results} shows the results for real world tracking. We evaluate on 32 motions ranging between 5 to 15 seconds each. For each motion, we rerun the evaluation 3 times for the PPO and QTOP checkpoints which are trained for $10^3$ GPU hours respectively on our $24$-hour training dataset and average R-MPKPE and joint MAE respectively. We note that joint MAE is computed over all joints where R-MPKPE is computed only over keybodies which are used in the tracking reward during training. Overall, we see a significant difference in both metrics. QTOP is around 8\% better in R-MPKPE and slightly less so in joint MAE, but the standard errors are non-overlapping. This suggests that the superiority we already observed in simulation translates to the real world. 

\section{Discussion}

We introduce QTOP, a motion tracking algorithm that consistently improves over strong on- and off-policy baselines in simulation and transfers these gains to real hardware. Our scaling analysis further shows that, within the measured regime of \(4\)--\(24\) dataset hours and up to approximately \(1,300\) GPU-hours, tracking remains primarily compute-limited, as increasing optimization compute generally yields larger gains than adding reference data. The main limitation of our work is that, due to limited resources, our experimental regime is one magnitude smaller in compute than state-of-the-art trackers. We therefore cannot yet determine whether the fitted relationship or QTOP's advantages persist at substantially larger data and compute scales. In particular, the importance of dataset size may increase once policies receive sufficient compute to exploit larger and more diverse motion corpora. Rather than advocating for scaling data and compute indiscriminately, our results motivate greater attention to algorithms that handle the continuous multi-task structure of motion tracking more efficiently. We view QTOP and our compute-data analysis as a starting point for developing such methods and for studying scaling behavior at larger regimes.

%Limitations: The main limitation is that we don't scale up to the same amount of compute and data that the large scale tracking works do due to resource limitations. Whether QTOP scales from 1000 GPU hours to 10,000 is future work. Also, we are quite sure that with more data, the influence of data in the scaling law might change. But nonetheless we want to emphasize that even with small data and 1000s of GPU hours you are still better off doubling your compute than your data. So in combination, neither are we building the best tracker nor are we proposing a scaling law that holds for much higher compute magnitudes but we propose a recipe that lets you be more efficient when training trackers on smaller compute and show that dataset size is not so important to get to something decent.

%Conclusions: In conclusion we propose QTOP which significantly beats state of the art RL algorithms in motion tracking. Also our paper gives critical insights in what matters really when scaling (data vs compute). We also hope that this is a starting point for further research into RL algorithms that scale well while handling the continuous multitask nature of motion tracking well and QTOP is a starting point to build upon. 

\section*{Acknowledgements}
Generative AI tools were used for coding assistance (GPT 5.5, GPT 5.6-sol), literature search (GPT 5.5), language editing (GPT 5.5), and as conversational aids for brainstorming scientific hypotheses (GPT 5.5). We note that all hypotheses were evaluated and selected on our own by conducting the experiments, interpreting the results and contextualizing them within the literature independently.


%%%%%%%%%%%%%%%%%%%%%%%%%%%%%%%%%%%%%%%%%%%%%%%%%%%%%%%%%%%%%%%%%%%%%%%%%%%%%%%%

%%%%%%%%%%%%%%%%%%%%%%%%%%%%%%%%%%%%%%%%%%%%%%%%%%%%%%%%%%%%%%%%%%%%%%%%%%%%%%%%

\bibliographystyle{IEEEtran}
\bibliography{biblio}

\end{document}